\documentclass{article}

\PassOptionsToPackage{numbers,sort&compress}{natbib}
\usepackage[preprint]{neurips_2026}

\usepackage[utf8]{inputenc}
\usepackage[T1]{fontenc}
\usepackage{hyperref,url}
\usepackage{booktabs}
\usepackage{longtable}
\usepackage{amsmath,amssymb,amsfonts,amsthm,mathtools}
\usepackage{microtype}

\usepackage{xcolor,graphicx,tikz}
\usetikzlibrary{arrows.meta,positioning}

\newtheorem{theorem}{Theorem}
\newtheorem{proposition}[theorem]{Proposition}
\newtheorem{corollary}[theorem]{Corollary}
\newtheorem{lemma}[theorem]{Lemma}
\theoremstyle{remark}

\newcommand{\R}{\mathbb{R}}
\newcommand{\E}{\mathbb{E}}
\newcommand{\AGP}{\mathrm{AGP}}
\newcommand{\dAGP}{\mathrm{dAGP}}
\newcommand{\cC}{\mathcal{C}}
\newcommand{\cP}{\mathcal{P}}
\newcommand{\norm}[1]{\left\lVert #1\right\rVert}

\usepackage{cleveref}
\crefname{section}{§}{§§}\crefname{subsection}{§}{§§}

\usepackage{enumitem}
\setlist[itemize,1]{leftmargin=1em} 
\setlist[enumerate,1]{leftmargin=1em} 

\title{Distributional Algebro-Geometric Persistence (dAGP):
\\
Wasserstein-Stable Polynomial-Relation Signatures}
\workshoptitle{Bridging Optimal Transport, Learning and Structured Data: Toward Geometric Distributional Learning}

\author{Edisy Kin Wai Chan \qquad\quad Andersen Ang 
\\
School of Electronics and Computer Science, \,
University of Southampton, \,
United Kingdom 
}

\begin{document}
\maketitle

\begin{abstract}
A circle and a rounded square are indistinguishable to homology: both are simple closed curves.
They are not indistinguishable algebraically, since their defining equations have degree $2$ and degree $4$.
We study a representation of point-cloud distributions that targets this algebraic structure.
For a probability measure $\mu$ on $\Omega\!=\![-1,1]^d$ and an orthonormal polynomial feature map $\phi_m$ of degree $m$, the moment matrix $G_m(\mu)\!=\!\E_{X\sim\mu}[\phi_m(X)\phi_m(X)^\top]$ satisfies $c^\top G_m(\mu)c\!=\!\E_\mu[f_c(X)^2]$, so coefficient directions associated with its small eigenvalues represent low-degree polynomials that nearly vanish on $\mu$.
Counting these near-relation directions across degree $m$ and the tolerance $\varepsilon$ gives a two-parameter integer signature.
We call this representation \emph{distributional algebro-geometric persistence} (dAGP), the population counterpart of the point-cloud algebro-geometric persistence (AGP) construction.
For empirical measures, dAGP is the sample-normalised counterpart of point-cloud AGP.
We also show that $\mu\!\mapsto\! G_m(\mu)$ is Lipschitz in the $1$-Wasserstein distance, so its ordered eigenvalues vary Lipschitz-continuously under transport of mass, while its threshold counts remain unchanged under perturbations satisfying a spectral-margin condition.
We give an explicit support-localised Lipschitz constant and identify the regime in which the resulting guarantee is informative.

Experimentally, the population noise-floor prediction agrees with the finite-sample mean to $0.1\%$, within sampling error.
At a selected tolerance, relation onset orders the conic, quartic, and non-algebraic closed curves, while their count profiles distinguish them from the two-dimensional control.
Calibration identifies an empirical tolerance range. A $100$-repetition sensitivity study quantifies onset recovery and shows that the count grid and two continuous spectral baselines nearly saturate the four-family discrimination task.
\end{abstract}

\section{Introduction}\label{sec:intro}
Geometric learning exploits structure that generic Euclidean representations obscure: symmetries, manifolds, graphs, and other non-Euclidean structure~\citep{bronstein2021geometric}.
A complementary viewpoint represents observations as probability distributions, using kernel mean embeddings~\citep{smola2007hilbert,muandet2017kernel} or geometry-aware discrepancies such as optimal transport (OT)~\citep{santambrogio2015optimal,peyre2019computational}.
We ask whether \emph{approximate polynomial relations} can serve as a third kind of structured representation of a point-cloud distribution.

Approximate polynomial relations encode information distinct from topology.
Homological summaries record connectivity: how many components, loops, and voids a space has~\citep{edelsbrunner2010computational,cohensteiner2007stability}.
Algebraic summaries describe the complexity of equations that define the space.
A circle $x^2\!+\!y^2\!=\!r^2$ and a quartic superellipse $x^4\!+\!y^4\!=\!r^4$ have identical low-dimensional homology but different minimal defining degrees.
Conversely, two conics can differ sharply in homology.
Estimating such algebraic invariants (degree, dimension, defining equations) from finite samples is the subject of a substantial literature~\citep{breiding2018learning,vidal2005gpca,yang2005hilbert}, and approximate vanishing-ideal methods recover near-vanishing polynomials from noisy data using evaluation matrices and spectral
thresholding~\citep{abbott2008stable,heldt2009approximate,livni2013vanishing,kera2018approximate,kera2020gradient}.
Our broader algebro-geometric persistence (AGP) framework\footnote{AGP is an unpublished ongoing project.
\cref{sec:signature} restates all AGP components used here: the approximate-vanishing cone, its spectral index, and degree--tolerance nesting.} organises such approximate relations across two axes, polynomial degree and tolerance.
The construction is \emph{persistence-style} instead of persistent homology: it tracks graded algebraic complexity, not the birth and death of homology classes.
Appendix~\ref{app:notation} collects the principal dAGP terminology and notation.
Other notation is defined where it first appears.

We introduce a population extension, \emph{distributional} AGP (dAGP).
Treating a point cloud as an empirical measure separates the population object from its finite-sample realisation. 
This formulation enables a direct stability analysis under transport of mass.
Our contributions are:
\begin{itemize}\itemsep0pt
\item a self-contained polynomial moment-matrix formulation of the degree--tolerance signature (\cref{sec:signature}), with a worked example that clarifies the interpretation of each count, and an exact correspondence with the point-cloud construction (Proposition~\ref{prop:empirical});
\item a $W_1$ Lipschitz bound for the moment operator and its ordered eigenvalues, together with a spectral-margin condition guaranteeing invariance of the threshold counts (\cref{sec:stability}).
Numerically, the \emph{localised} constant is $2.5$--$19\times$ smaller than the cube-wide value over $m=2,\dots,6$.
We quantify the regime in which the guarantee is informative;
\item five experimental studies (\cref{sec:experiments}): perturbation response, controlled relation onset and its noise floor, tolerance calibration, sample-size sensitivity with spectral comparisons, and a multi-shape diagnostic.
\end{itemize}

\section{The dAGP signature}\label{sec:signature}

\paragraph{Polynomial features.}
Let $\Omega\!=\![-1,1]^d$. 
Let $S_{\le m}\!=\!\R[x_1,\dots,x_d]_{\le m}$ be the vector space of polynomials of total degree at most $m$.
Its dimension is $M_m\!=\!\binom{d+m}{m}$.
Fix a nested basis $\{p_\alpha:|\alpha|\!\le\! m\}$ of $S_{\le m}$ that is orthonormal in $L^2(\Omega,dx)$.
In our experiments $p_\alpha(x)\!=\!\prod_{j=1}^d\ell_{\alpha_j}(x_j)$, where $\ell_k(t)\!=\!\sqrt{\frac{2k+1}{2}}P_k(t)$ is the $L^2([-1,1],dt)$-normalised Legendre polynomial~\citep{dunkl2014orthogonal}.
Write
\[
\phi_m(x)=\bigl(p_\alpha(x)\bigr)_{|\alpha|\le m}\in\R^{M_m},
\]
so that a coefficient vector $c\!\in\!\R^{M_m}$ represents the polynomial $f_c=\!\sum_\alpha c_\alpha p_\alpha$ and $\phi_m(x)^\top c\!=\!f_c(x)$.

\paragraph{The moment matrix.}
For $\mu\in\cP(\Omega)$ define
\begin{equation}\label{eq:moment}
G_m(\mu)\coloneqq\int_\Omega \phi_m(x)\phi_m(x)^\top d\mu(x),
\quad\text{so that}\quad
c^\top G_m(\mu)\,c=\int f_c(x)^2 d\mu(x)=\norm{f_c}^2_{L^2(\mu)}.
\end{equation}
\eqref{eq:moment} links the spectrum of $G_m(\mu)$ to approximate polynomial relations.
A unit-norm coefficient vector $c$ with small $c^\top G_m(\mu)c$ represents a polynomial in $S_{\le m}$ that is close to zero in mean square with respect to $\mu$: an \emph{approximate relation} satisfied by the distribution.
Hence the small eigenvalues of $G_m(\mu)$, not the large ones, encode the signal.
Fixing a tolerance $\varepsilon>0$, set
\begin{equation}\label{eq:signature}
\cC^{\dAGP}_{m,\varepsilon}(\mu)\coloneqq\{c:\norm{f_c}_{L^2(\mu)}\le\varepsilon\norm{c}_2\},
\qquad
\begin{aligned}
r^{\dAGP}_{m,\varepsilon}(\mu)&\coloneqq\#\{k:\lambda_k(G_m(\mu))\le\varepsilon^2\},\\
H^{\dAGP}_{m,\varepsilon}(\mu)&\coloneqq M_m-r^{\dAGP}_{m,\varepsilon}(\mu).
\end{aligned}
\end{equation}
For fixed $m$, the map $\varepsilon\mapsto r^{\dAGP}_{m,\varepsilon}(\mu)$ is the eigenvalue counting function of $G_m(\mu)$ and hence contains the same information as its ordered spectrum.
The additional structure comes from organising these counting functions consistently across polynomial degree and interpreting their onsets algebraically.

$G_m(\mu)$ is the degree-$m$ moment matrix, equivalently the Gram matrix of the chosen polynomial features in $L^2(\mu)$ \citep{curto2005truncated,pauwels2021data}.
When $G_m(\mu)$ is nonsingular, its inverse defines the associated Christoffel--Darboux kernel~\citep{lasserre2022book}.

\paragraph{What the two counts mean.}
The set $\cC^{\dAGP}_{m,\varepsilon}$ is a quadratic cone, not a subspace, but its largest linear subspace has dimension exactly $r^{\dAGP}_{m,\varepsilon}$ (Lemma~\ref{lem:inertia}).
So $r^{\dAGP}_{m,\varepsilon}$ counts independent directions in coefficient space along which the distribution admits a near-relation,
and $H^{\dAGP}_{m,\varepsilon}$ counts the complementary directions: how many genuinely different polynomial functions of degree at most $m$ the distribution still distinguishes once its own constraints have been imposed.
At zero tolerance, let $V=\overline{\operatorname{supp}\mu}^{\,Z}$ be the Zariski closure of the support and let $I(V)$ be its vanishing ideal.
By \citet[Proposition~1]{pauwels2021data},
\[
H^{\dAGP}_{m,0}(\mu)
~=~
\operatorname{rank}G_m(\mu)
~=~
\dim\!\left( S_{\le m}/\big(I(V)\cap S_{\le m}\big) \right)
~=~
\operatorname{HF}_V(m),
\]
where $\operatorname{HF}_V$ is the affine Hilbert function of $V$.
For $\varepsilon>0$, we regard $H^{\dAGP}_{m,\varepsilon}$ as a tolerance-dependent effective analogue of the exact Hilbert function. 
We hence call it a \emph{Hilbert-style effective algebraic dimension}.
Hilbert functions have previously been used as descriptors of subspace arrangements~\citep{yang2005hilbert}.

\begin{figure}[h!]
\includegraphics[width=\linewidth]{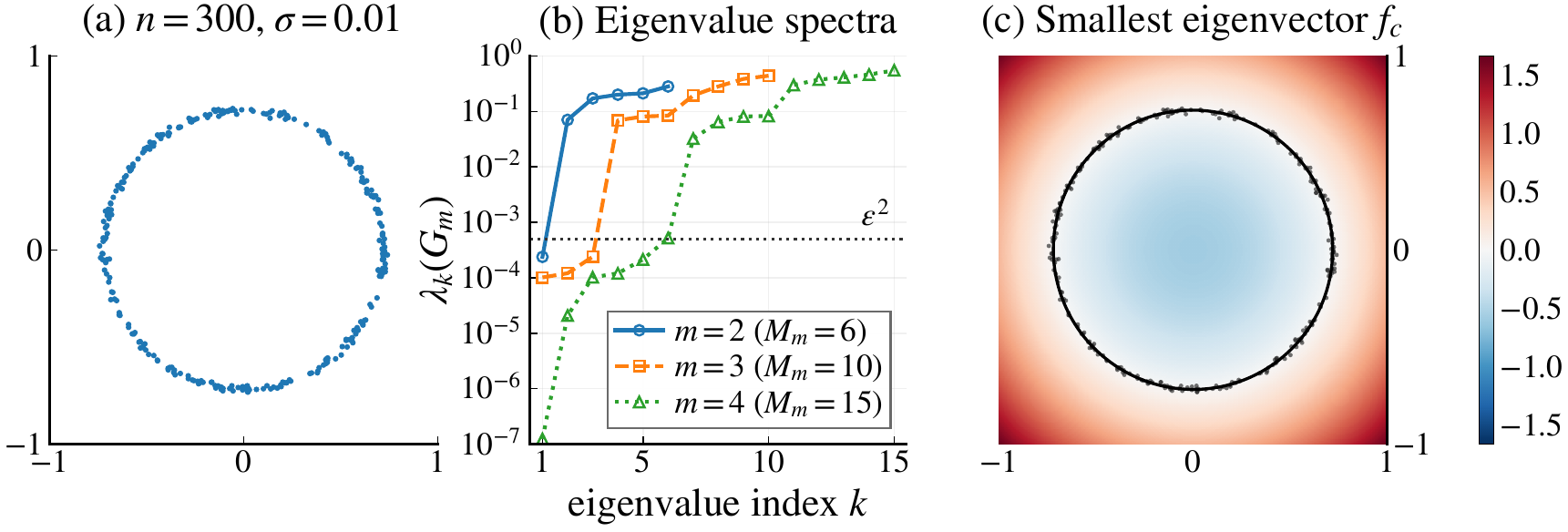}
\vspace{-5mm}
\caption{The dAGP construction for one example.
(a) $n\!=\!300$ points on a radius-$0.72$ circle with noise $\sigma\!=\!0.01$, inside $\Omega\!=\![-1,1]^2$.
(b) The spectra of $G_m(\widehat\mu_Z)$ give counts $1,3,5$ at $m\!=\!2,3,4$; the sixth predicted $m\!=\!4$ eigenvalue is $5.19\times10^{-4}$, just above the threshold.
The local gap is the margin relevant to Corollary~\ref{cor:threshold}.
(c) The eigenvector for the smallest eigenvalue at $m\!=\!2$, drawn as a function on $\Omega$; its zero set is the black curve.
Its coefficients agree with those of $x^2+y^2-r^2$ to within $0.4\%$.
So the smallest-eigenvalue direction recovers the defining equation.}
\label{fig:worked}
\end{figure}

\paragraph{A worked example.}
Take $\mu$ uniform on the circle of radius $r\!=\!0.72$ in $\Omega\!=\![-1,1]^2$, with isotropic Gaussian noise of scale $\sigma\!=\!0.01$.
At degree $m\!=\!2$ there are $M_2\!=\!6$ basis polynomials.
Exactly one combination, $g(x,y)\!=\!x^2+y^2-r^2$, vanishes on the noiseless circle, so we expect $r^{\dAGP}_{2,\varepsilon}\!=\!1$ and $H^{\dAGP}_{2,\varepsilon}\!=\!5$.
At $m\!=\!3$ the multiples $\{1,x,y\}\cdot g$ all vanish, predicting $r\!=\!3$; at $m\!=\!4$ the multiples by $\{1,x,y,x^2,xy,y^2\}$ predict $r\!=\!6$.
Measured counts over 20 samples are $1$, $3$, and $5.1\pm0.3$ (\cref{sec:onset}).
At $m\!=\!4$, 18 out of the 20 draws count five and the remaining two count six at $\varepsilon^2\!=\!5\times10^{-4}$, whereas all 20 count six at $10^{-3}$, still inside the calibration range of \cref{sec:calibration}: the lower threshold cuts through the noise-lifted band.
Fig.~\ref{fig:worked} shows the whole pipeline for one draw.

The noise also predicts \emph{how small} the corresponding eigenvalue should be.
Writing $z\!=\!z_0+e$ with $|z_0|\!=\!r$ and $e\sim\mathcal N(0,\sigma^2I_2)$ gives $g(z)\!=\!2\langle z_0,e\rangle+|e|^2$ and hence $\E[g^2]\!=\!4r^2\sigma^2+8\sigma^4$.
In the normalised Legendre basis $g$ has coefficients $c_{(2,0)}\!=\!c_{(0,2)}\!=\!4/(3\sqrt5)$ and $c_{(0,0)}\!=\!\tfrac43-2r^2$, so
\begin{equation}\label{eq:floor}
\lambda_{\min}(G_2)
~\le~
\frac{4r^2\sigma^2+8\sigma^4}{2c_{(2,0)}^2+c_{(0,0)}^2}
~=~
2.596\times10^{-4}.
\end{equation}
The right-hand side is a population Rayleigh quotient.
Across 20 independent empirical measures, the mean $\lambda_{\min}(G_2)$ is $2.598\times10^{-4}$ with sample standard deviation $2.1\times10^{-5}$.
The difference from the population quotient is well below one standard error.
The $0.1\%$ difference is consistent with finite-sample variation.
Eq.~\eqref{eq:floor} is useful beyond the sanity check: the quotient provides an upper bound on $\lambda_{\min}(G_2(\mu))$ for the noisy population measure.
Hence, choosing $\varepsilon^2$ at least as large as the displayed quantity is sufficient to count at least one near-relation.
The bound is empirically nearly tight in this symmetric example, so we use it as a practical noise-floor estimate to calibrate $\varepsilon$ in \cref{sec:calibration}.

\paragraph{Monotonicity.}
Zero-padding $S_{\le m}\!\hookrightarrow\! S_{\le m'}$ leaves $f_c$ and $\norm{c}_2$ unchanged, so the cones are nested in degree and also nested in $\varepsilon$.
Since $G_m(\mu)$ is a principal submatrix of $G_{m+1}(\mu)$, Cauchy interlacing gives $H^{\dAGP}_{m,\varepsilon}\!\le\! H^{\dAGP}_{m+1,\varepsilon}$ (Appendix~\ref{app:nesting}).
So $(m,\varepsilon)\!\mapsto\! r^{\dAGP}_{m,\varepsilon}(\mu)$ is a monotone integer-valued function of two parameters: a bifiltration-indexed invariant.
Our construction is persistence-style.\footnote{
\emph{Persistence-style} refers only to this two-parameter nesting, not to a persistence module, barcode decomposition, or interleaving theory~\citep{chazal2016structure,lesnick2015interleaving}.}

\paragraph{Relation to the point-cloud construction.}
For $Z\!=\!\{z_i\}_{i=1}^n\!\subset\!\Omega$ let $\Phi_m(Z)\!\in\!\R^{n\times M_m}$ have rows $\phi_m(z_i)^\top$, so $\Phi_m(Z)c$ is the evaluation vector of $f_c$.
Point-cloud AGP uses the approximate-vanishing cone $\cC^{\AGP}_{m,\tau}(Z)\!=\!\{c:\norm{\Phi_m(Z)c}_2\!\le\!\tau\norm{c}_2\}$ with index $r^{\AGP}_{m,\tau}(Z)\!=\!\#\{k:\sigma_k(\Phi_m(Z))\!\le\!\tau\}$.\footnote{Singular values are counted with the convention that $\Phi_m(Z)$ has $M_m$ of them; when $n\!<\!M_m$ the list is padded with $M_m\!-\!n$ zeros, which is what the maximal-subspace identity requires.}

\begin{proposition}[Empirical correspondence]\label{prop:empirical}
For $\widehat\mu_Z\!=\!n^{-1}\sum_{i=1}^n\delta_{z_i}$,
\begin{equation}\label{eq:correspondence}
G_m(\widehat\mu_Z)
\!=\!
\tfrac1n\Phi_m(Z)^\top\Phi_m(Z),\qquad
\cC^{\dAGP}_{m,\varepsilon}(\widehat\mu_Z)
\!=\!
\cC^{\AGP}_{m,\sqrt n\,\varepsilon}(Z).
\end{equation}
\end{proposition}
dAGP is thus the sample-normalised population counterpart of point-cloud AGP, and the two tolerance scales differ by $\sqrt n$.
We report every tolerance in population units $\varepsilon$, so that quantities computed at different sample sizes are comparable.

\section{Wasserstein stability}\label{sec:stability}
Because $\phi_m$ is a fixed polynomial map on a compact set, the constants
\begin{equation}\label{eq:regularity}
B_{m,d}\coloneqq\sup_{x\in\Omega}\norm{\phi_m(x)}_2,\qquad
L_{m,d}\coloneqq\sup_{x\in\Omega}\norm{D\phi_m(x)}_2
\end{equation}
are finite.

\begin{theorem}[Wasserstein stability]\label{thm:w1}
Let $K\!\subseteq\!\Omega$ be convex with $\mathrm{supp}\,\mu\cup\mathrm{supp}\,\nu\!\subseteq\! K$, and let $B_{m,d}(K),L_{m,d}(K)$ denote the suprema in \eqref{eq:regularity} taken over $K$.
Then
\begin{equation}\label{eq:w1bound}
\norm{G_m(\mu)-G_m(\nu)}_2\le 2B_{m,d}(K)L_{m,d}(K)\,W_1(\mu,\nu),
\end{equation}
and every ordered eigenvalue satisfies $|\lambda_k(G_m(\mu)) \!-\!\lambda_k(G_m(\nu))|\!\le\! 2B_{m,d}(K)L_{m,d}(K)\,W_1(\mu,\nu)$.
\end{theorem}

\begin{corollary}[Threshold stability]\label{cor:threshold}
Let $\Delta\!=\!2B_{m,d}(K)L_{m,d}(K)W_1(\mu,\nu)$.
If $G_m(\mu)$ has no eigenvalue in $[\varepsilon^2-\Delta,\varepsilon^2+\Delta]$ then $r^{\dAGP}_{m,\varepsilon}(\mu)\!=\!r^{\dAGP}_{m,\varepsilon}(\nu)$, and likewise for $H^{\dAGP}_{m,\varepsilon}$.
\end{corollary}
A short proof is given in Appendix~\ref{app:proofs}.
The map $x\!\mapsto\!\phi_m(x)\phi_m(x)^\top$ is Lipschitz on $K$ with constant $2B_{m,d}(K)L_{m,d}(K)$.
Integration over a coupling and taking the infimum gives \eqref{eq:w1bound}, and Weyl's inequality supplies the eigenvalue bound.
The contribution here is the explicit constant and its quantitative interpretation.
What matters for use is the size of the constant, which the statement above lets one control by shrinking $K$, and this is the reason for stating the localised form: the suprema in \eqref{eq:regularity} are attained at the corners of the cube, which measures supported near the origin never visit.

\begin{table}[t]
\centering\small
\caption{Numerical grid estimates of the feature-map constants for tensor-product Legendre features on $[-1,1]^2$, computed on a $401^2$ grid.
The last column restricts the computation to $K\!=\!B_{0.9}$, the ball of radius $0.9$, which contains the empirical circle sample used for the illustrative stability calculation in \cref{sec:stability}.
Localising reduces the estimated product $2B_{m,2}L_{m,2}$ by $7\times$ at $m\!=\!4$ and $19\times$ at $m\!=\!6$, and changes the log--log slope fitted over the displayed degrees $m\geq3$ from approximately $4.6$ to $2.2$.}
\label{tab:constants}
\setlength{\tabcolsep}{5pt}
\begin{tabular}{@{}rrrrrr@{}}
& & & & \multicolumn{2}{c}{$2B_{m,2}L_{m,2}$}\\
\cmidrule(l){5-6}
$m$ & $M_m$ & $B_{m,2}$ & $L_{m,2}$ & $K\!=\!\Omega$ & $K\!=\!B_{0.9}$\\
\hline\hline\noalign{\vskip 2pt}
1 & 3 & 1.32 & 0.87 & 2.3 & 1.6\\
2 & 6 & 2.55 & 4.06 & 20.7 & 8.4\\
3 & 10 & 4.18 & 11.81 & 98.8 & 23.7\\
4 & 15 & 6.23 & 27.10 & 337.4 & 48.3\\
5 & 21 & 8.68 & 53.65 & 930.8 & 80.9\\
6 & 28 & 11.53 & 95.95 & 2213.2 & 118.9\\
8 & 45 & 18.47 & 249.44 & 9215.9 & 206.8
\end{tabular}
\vspace{-3mm}
\end{table}

\paragraph{When does Corollary~\ref{cor:threshold} bite?}
For perturbations $\nu$ also supported in $K$, the guarantee is informative when $\Delta$ is smaller than the spectral gap around $\varepsilon^2$; its certified radius is $W_1\!<\!\mathrm{gap}/(2B_{m,d}(K)L_{m,d}(K))$.
For the circle draw in Fig.~\ref{fig:worked} at $\varepsilon^2\!=\!5\times10^{-4}$, the gaps at $m\!=\!2,4$ are $2.6\!\times\!10^{-4}$ and $1.9\!\times\!10^{-5}$, giving indicative localised radii $3.1\!\times\!10^{-5}$ and $3.9\!\times\!10^{-7}$ (and $2.5$--$7\!\times\!$ smaller cube-wide values).
At $\varepsilon^2\!=\!10^{-3}$, the sixth $m\!=\!4$ eigenvalue lies approximately $4.8\!\times\!10^{-4}$ below the threshold; hence the gap and localised radius are at most $4.8\!\times\!10^{-4}$ and $1.0\times10^{-5}$, the latter still over two orders below the smallest tested $W_1\!=\!6.0\!\times\!10^{-3}$.
Thus Theorem~\ref{thm:w1} gives Lipschitz continuity of the moment operator and its ordered eigenvalues.
Corollary~\ref{cor:threshold} makes threshold counts locally constant under a computable margin.
The resulting threshold guarantee is not tight at the informative tolerances.
Sharper guarantees may require relative perturbation bounds or constants adapted to the measure rather than only to its support.

\paragraph{Convergence and other metrics.}
Taking $\nu\!=\!\widehat\mu_n$ in Theorem~\ref{thm:w1} shows that the empirical moment matrix converges almost surely to its population counterpart.
Empirical $W_1$ rates also yield an explicit expectation bound.
The same operator is controlled by the MMD of a lifted polynomial kernel, but fixed-degree moments are not injective, so none of these bounds is bi-Lipschitz.
The calculations and precise qualifications are in Appendix~\ref{app:metrics}.

\section{Experiments}\label{sec:experiments}
All experiments use small synthetic two-dimensional datasets; they are designed to test specific claims rather than to demonstrate downstream performance.
Spectra are computed from the singular values of $\Phi_m/\sqrt n$ rather than by eigendecomposing $G_m$, which avoids squaring the condition number; at $m\!=\!6$ the reported eigenvalues are about five orders of magnitude above the double-precision floor, and accordingly, we use $m\!\le\!6$ in \cref{sec:perturbation}--\cref{sec:samplesize}.
Full specifications, seeds, and the constant computation are in Appendix~\ref{app:experiments}.

\subsection{Perturbation response}\label{sec:perturbation}
A clean empirical measure $\widehat\mu_0$ contains $n\!=\!160$ points on a radius-$0.72$ circle.
For each $\sigma\in\{0.005,0.01,0.02,0.04,0.07,0.10,0.14\}$, we generate $12$ independent isotropic Gaussian perturbations and project the perturbed points onto the cube.
We compute the exact $W_1$ between the clean and perturbed equal-weight empirical measures by linear assignment.
At $m\!=\!4$, we measure the operator displacement and the maximum absolute displacement among the ordered eigenvalues.

\begin{figure}[h!]
\includegraphics[width=\linewidth]{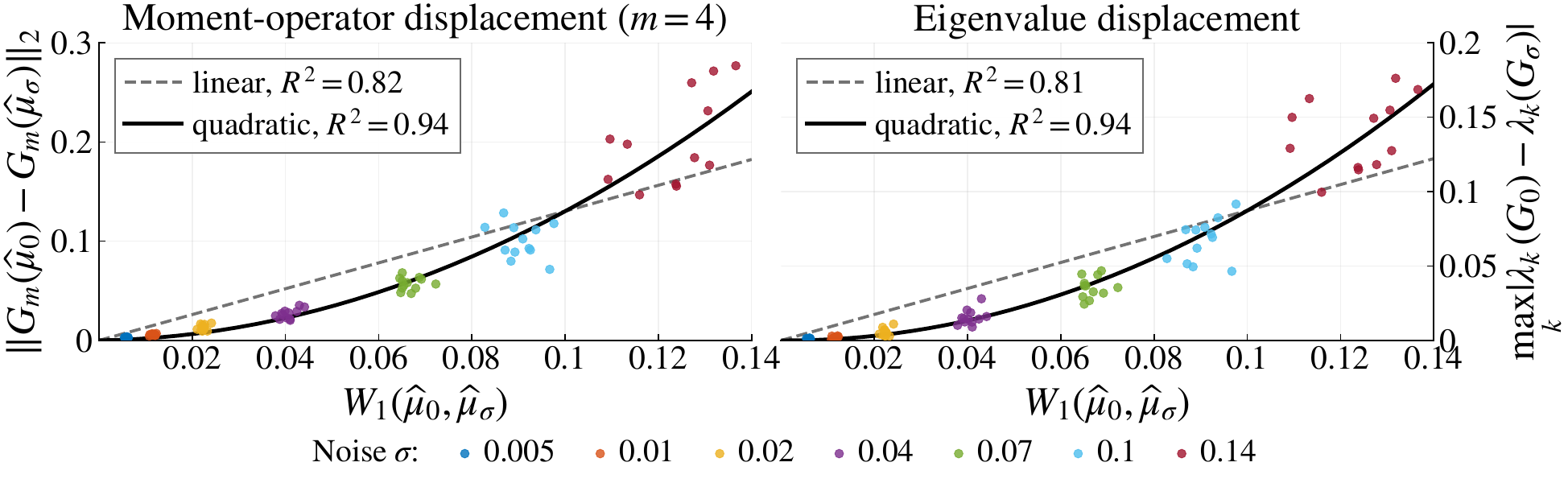}
\vspace{-5mm}
\caption{Perturbation response at $m\!=\!4$; colours denote noise $\sigma$ in both panels.
Both displacements grow smoothly with $W_1$; for the operator, $\|\Delta G\|_2/W_1$ rises monotonically from $0.47$ to $1.6$, showing upward curvature over the tested grid.
A least-squares line constrained through the origin gives $R^2\!=\!0.82$ and $0.81$ in the two panels.
A linear-plus-quadratic fit gives $R^2\!=\!0.94$ in both panels.
These raw-scale fits are descriptive and weighted towards the largest perturbations, not estimates of the worst-case constant in Theorem~\ref{thm:w1}; the cube-wide value in Table~\ref{tab:constants} is $337$.}
\label{fig:wasserstein}
\end{figure}

Fig.~\ref{fig:wasserstein} shows the response.
Eigenvalue displacement stays below operator displacement, as Weyl's inequality requires, so this result checks the implementation rather than the theory.
The ratio of operator displacement to $W_1$ rises monotonically from $0.47$ to $1.6$ across the noise grid, so a single proportionality constant does not describe the full range.
The upward curvature is compatible with centered perturbations: since $\E[e]\!=\!0$, a Taylor-expansion heuristic suggests an $O(\sigma^2)$ population bias plus an $O(\sigma/\sqrt n)$ empirical fluctuation, while $W_1\asymp\sigma$.
The fits in Fig.~\ref{fig:wasserstein} are illustrative only and do not establish an asymptotic scaling law.
The cube-wide constant in the worst-case bound from Theorem~\ref{thm:w1} is roughly $210$ times the observed ratio at the largest noise level and more than $700$ times that ratio at the smallest.

\subsection{Relation onset}\label{sec:onset}
To test what the degree axis measures, we compare four planar families with $n\!=\!300$ over $20$ repetitions (noise $\sigma\!=\!0.01$ on the three curves only): the circle $x^2+y^2\!=\!r^2$, whose defining relation lies in $S_{\le2}$; the quartic superellipse $x^4+y^4\!=\!r^4$, whose displayed relation first lies in $S_{\le4}$ (both with $r\!=\!0.72$); a smooth simple closed curve given in polar form by $r(\theta)\!=\!r_0e^{a\cos k\theta}$ with $k\!=\!5$, which is one-dimensional but not an algebraic curve; and a uniform sample from $[-0.72,0.72]^2$.
The transcendental curve is the key control.
A comparison with the two-dimensional uniform square alone confounds intrinsic dimension with the absence of low-degree relations.
The transcendental curve holds intrinsic dimension fixed while varying algebraic approximability.

\begin{figure}[h!]
\includegraphics[width=\linewidth]{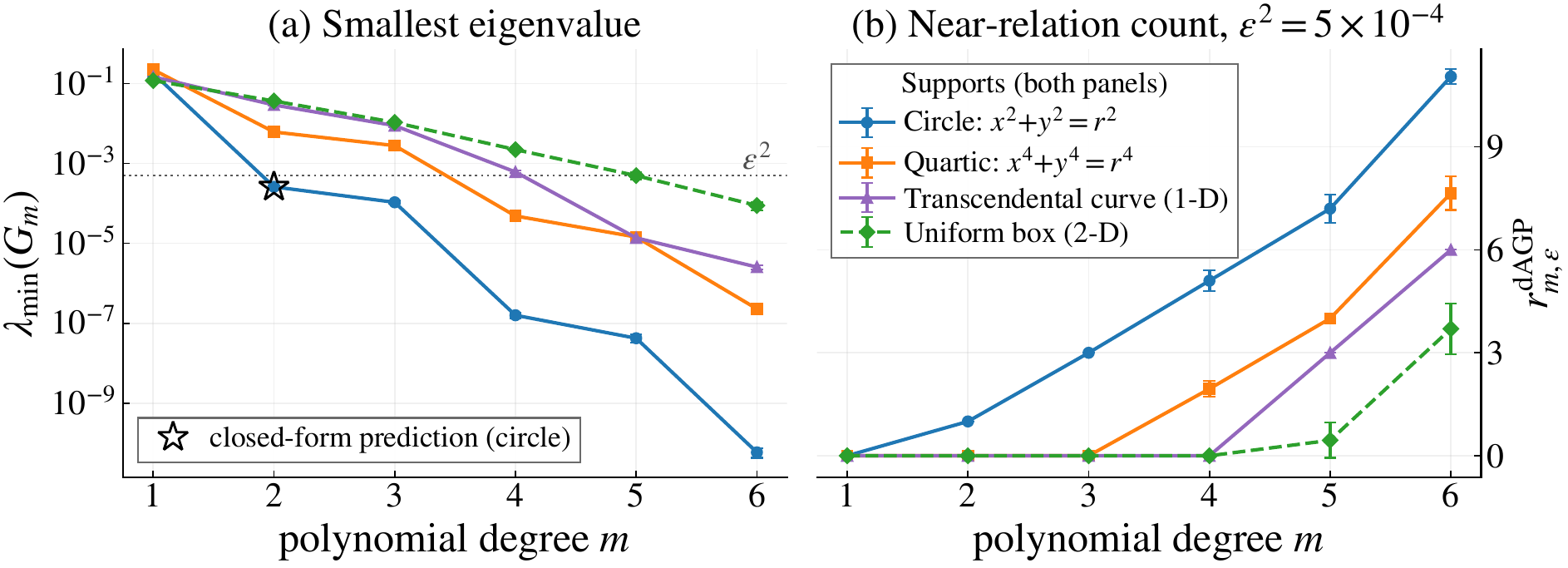}
\vspace{-5mm}
\caption{Relation onset over $20$ samples; error bars are one standard deviation.
(a) $\lambda_{\min}(G_m)$, with the closed-form prediction \eqref{eq:floor} for the circle at $m\!=\!2$ marked by a star and the calibrated tolerance shown as a dotted line.
(b) The near-relation count $r^{\dAGP}_{m,\varepsilon}$, which is the defining quantity of the dAGP signature.
At $n\!=\!300$, the first nonzero mean counts occur at degrees $2$, $4$, $5$, and $5$; Fig.~\ref{fig:samplesize} examines sample-wise reliability and the large-sample onsets.}
\label{fig:onset}
\end{figure}

Fig.~\ref{fig:onset} reports both $\lambda_{\min}(G_m)$ and the near-relation count, which is the object actually defined in \eqref{eq:signature}.
The count provides clearer separation in this experiment and is zero for every family below onset.
For the circle, fixed-threshold counts match the noiseless multiplicities at $m\!=\!2,3$ but undercount them for $m\!\ge\!4$.
At $\varepsilon^2\!=\!10^{-3}$ all draws recover the $m\!=\!4$ multiplicity; recovering $1,3,6,10,15$ through $m\!=\!6$ requires $2\!\times\!10^{-3}$, the upper edge of the calibration range (\cref{sec:calibration}).
We retain $5\!\times\!10^{-4}$, about twice the predicted noise floor, because looser thresholds recover more circle multiples but tie the quartic and transcendental onsets (\cref{sec:calibration}).
This trade-off motivates using a tolerance grid (\cref{sec:samplesize}) rather than a single operating point.
Selecting the grid relative to the noise level remains an open design question.
At the selected $\varepsilon^2\!=\!5\!\times\!10^{-4}$, the onsets $2\!<\!4\!<\!5$ order the circle, quartic, and transcendental curve by low-degree algebraic complexity (approximability for the latter).
The two-dimensional control is last at the large-sample reference (degree $6$, Fig.~\ref{fig:samplesize}), though at $n\!=\!300$ it ties the transcendental curve under the first-nonzero-mean summary; Fig.~\ref{fig:onset}(b) nevertheless separates their full profiles.
Fig.~\ref{fig:calibration}(a) instead averages sample-wise onset degrees across separate draws, explaining its noninteger curves and why they need not coincide with that summary.
The $\lambda_{\min}$ profile alone does not expose the degree-organised structure, since it records only the single smallest direction and falls smoothly for every family.

Under finite sampling and noise, onset degree measures tolerance-dependent low-degree algebraic approximability rather than exact minimal defining degree.
Finite samples and noise can generate approximate lower-degree relations.
Higher-degree multiples of an existing relation create additional small directions.
A curve that is well approximated by a conic can also register at low degree even when it is non-algebraic: with $k\!=\!2$ instead of $k\!=\!5$ above, the transcendental curve is close enough to an ellipse that its onset falls to degree $2$ or $3$ across samples.

\subsection{Tolerance calibration and complexity}\label{sec:calibration}
The separation in Fig.~\ref{fig:onset} depends on $\varepsilon$.
This dependence follows directly from the threshold construction: a threshold below the practical noise-floor estimate \eqref{eq:floor} may miss the defining relation, whereas a threshold far above it admits many spurious near-relations.

\begin{figure}[h!]
\includegraphics[width=\linewidth]{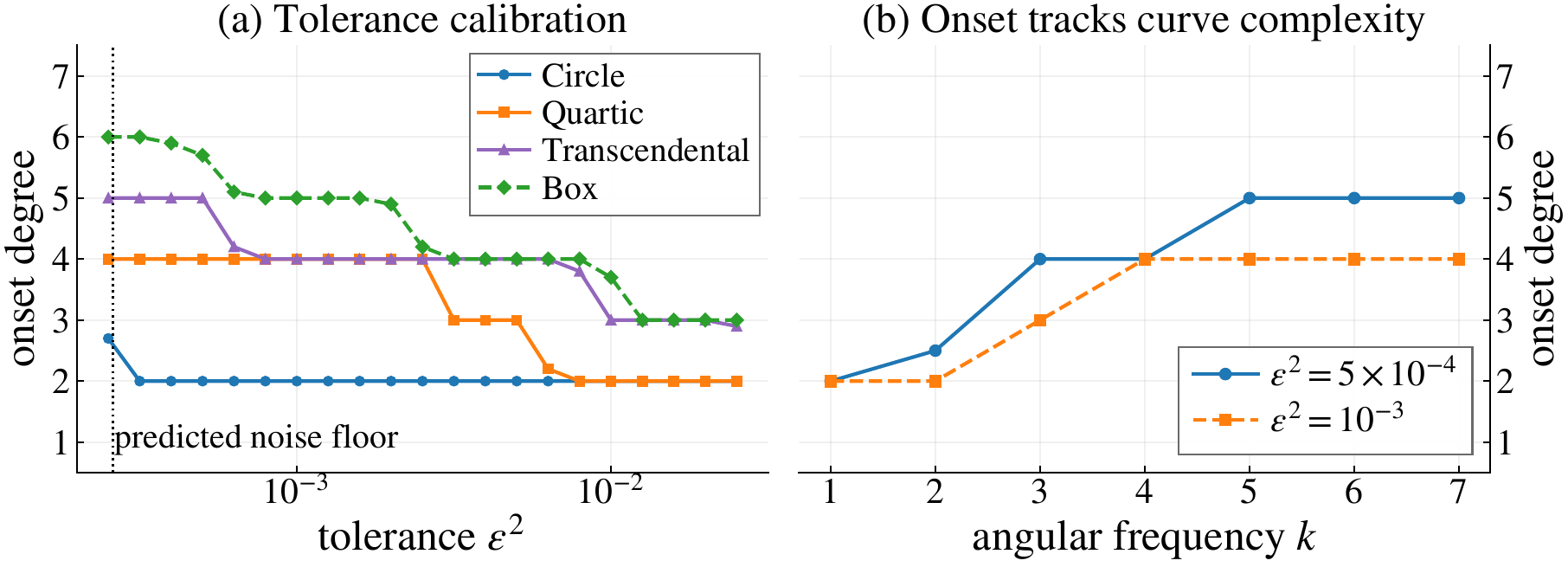}
\vspace{-5mm}
\caption{Mean sample-wise onset degree over 10 draws.
(a)~Versus squared tolerance~$\varepsilon^2$.
The quartic and transcendental curves share onset degree~$4$ from about $\varepsilon^2\!=\!7\times10^{-4}$ to the upper edge of the calibration range; looser tolerances eventually merge the quartic and circle onsets.
(b)~Versus angular frequency~$k$ for the family $r(\theta)\!=\!r_0e^{a\cos k\theta}$, at $\varepsilon^2\!=\!5\times10^{-4}$ and~$10^{-3}$.
Onset rises with~$k$, then saturates at degree~$5$ and~$4$ respectively: at this noise level, the sampled data from the tested curves exhibit near-relations at the corresponding degrees and tolerances.}
\label{fig:calibration}
\end{figure}

Fig.~\ref{fig:calibration}(a) suggests an empirical calibration range of roughly $3\!\times\!10^{-4}\lesssim\varepsilon^2\lesssim2\!\times\!10^{-3}$ above the practical noise floor.
In the plotted draws, the three curve families have distinct onsets at the selected $\varepsilon^2\!=\!5\!\times\!10^{-4}$; by about $7\!\times\!10^{-4}$, the quartic and transcendental curves both register degree~$4$. 
Individual draws begin to tie as $\varepsilon^2$ exceeds the transcendental curve's $\lambda_{\min}(G_4)$, $6.1\!\times\!10^{-4}$ on average (Fig.~\ref{fig:onset}(a)).
At $\varepsilon^2\!=\!10^{-3}$ and $2\!\times\!10^{-3}$, the two curves' counts at $m\!=\!5,6$ have disjoint ranges across the 20 draws of Fig.~\ref{fig:onset}, so their full count profiles still separate.
Fig.~\ref{fig:calibration}(b) shows that onset tracks the non-algebraic family's angular frequency before saturating.

\subsection{Sample-size sensitivity}\label{sec:samplesize}
\begin{figure}[h!]
\includegraphics[width=\linewidth]{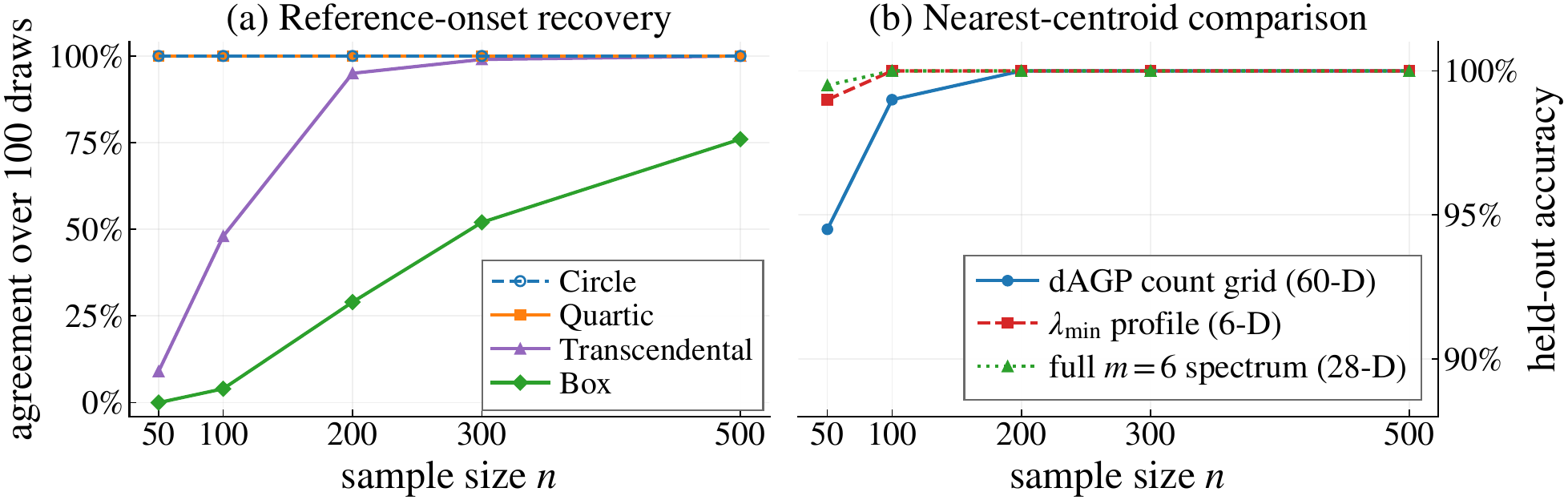}
\vspace{-5mm}
\caption{Finite-sample sensitivity over $100$ draws per support and $n$.
(a) Agreement with reference onsets $(2,4,5,6)$; the circle and quartic coincide at $100\%$ throughout.
(b) Nearest-centroid accuracy after training-fold standardisation for the $60$-coordinate degree--tolerance count grid, the six $\log\lambda_{\min}$ values, and the $28$ ordered log-eigenvalues at $m\!=\!6$.
The protocol is in Appendix~\ref{app:experiments}; the accompanying script writes every per-sample result.}
\label{fig:samplesize}
\end{figure}

For $n\in\{50,100,200,300,500\}$ we draw $100$ independent point clouds from each family in \cref{sec:onset}.
Fig.~\ref{fig:samplesize}(a) compares the fixed-threshold onset with a high-resolution Monte Carlo reference.
Circle and quartic recover their reference degrees on every draw, while the transcendental and box onsets converge more slowly because their relevant eigenvalues lie closer to the threshold.
Panel~(b) compares the three representations quantitatively.
The dAGP count-grid accuracy is $94.5\%$ at $n\!=\!50$, $99\%$ at $n\!=\!100$, and $100\%$ thereafter, while the $\lambda_{\min}$ degree profile and full $m\!=\!6$ spectrum reach $99$--$100\%$ already at $n\!=\!50$.
The task is saturated: all three representations achieve near-perfect discrimination, so this experiment cannot meaningfully rank them.
Distinguishing their finite-sample utility would require a harder matched task, for example degree-$4$ and degree-$6$ algebraic curves with the same intrinsic dimension, noise level, and sampling density.

\subsection{Multi-shape diagnostic}\label{sec:multishape}
We evaluate eight planar families ($n\!=\!250$, fixed seed) over $m\!\le\!10$ and a logarithmic grid of $\varepsilon$.
For this diagnostic only, we follow the point-cloud AGP preprocessing pipeline of centering, ridge whitening with $\lambda\!=\!10^{-3}$, and coordinatewise rescaling into $[-1,1]^2$; Theorem~\ref{thm:w1} does not cover this data-dependent preprocessing.
Accordingly, the other experiments omit it.

\begin{figure}[t]
\centering
\includegraphics[width=\linewidth]{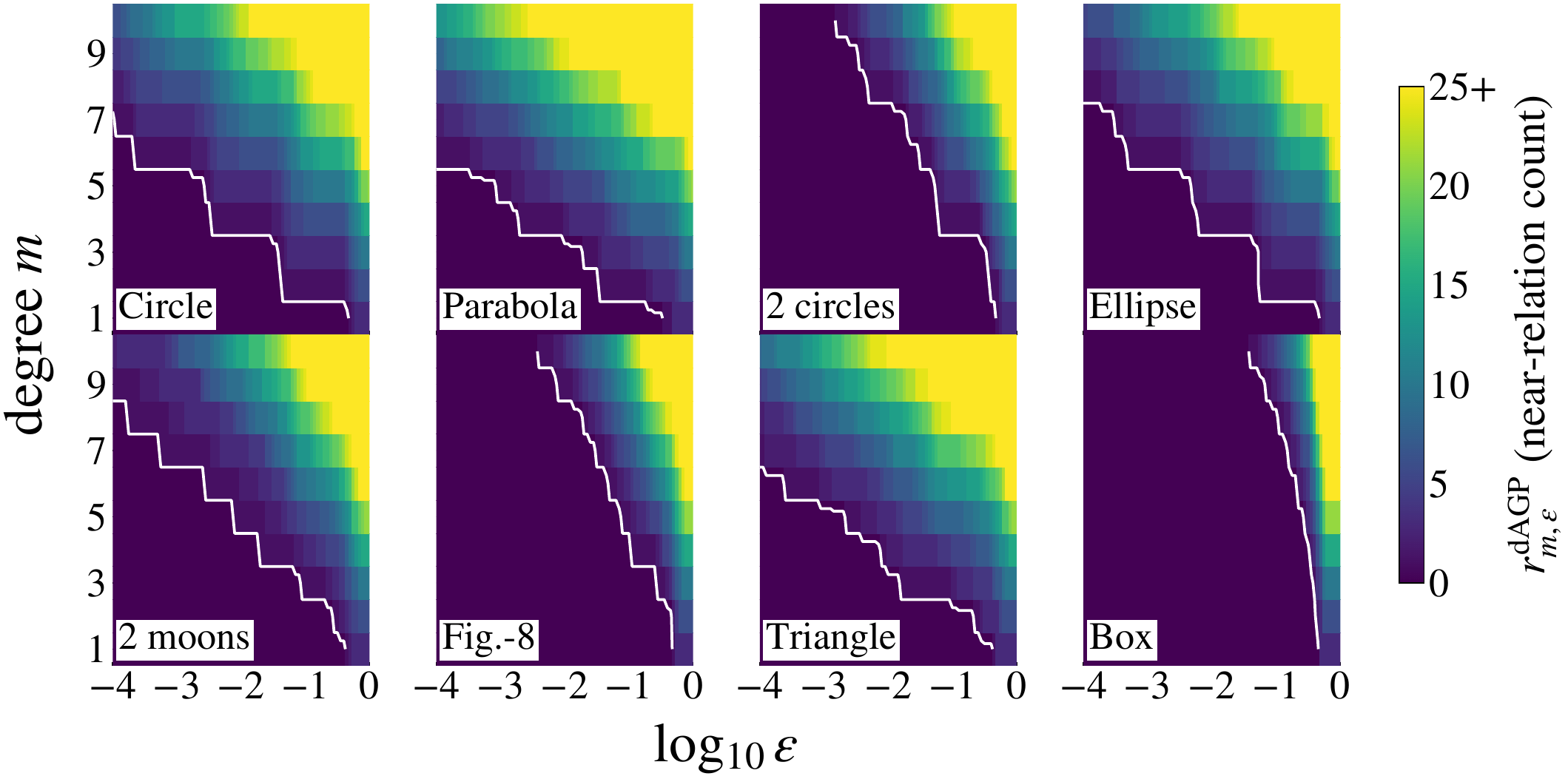}
\vspace{-5mm}
\caption{Near-relation counts $r^{\dAGP}_{m,\varepsilon}$ across the degree--tolerance grid, in population units.
The white contour marks the onset boundary $r^{\dAGP}_{m,\varepsilon}\!=\!1$.
Structured supports have onset boundaries that sweep to small $\varepsilon$ at low degree; the uniform-box control has a nearly vertical boundary at large $\varepsilon$, acquiring relations only when the tolerance is loose.
Circle and ellipse remain similar because whitening is applied here.}
\label{fig:heatmaps}
\vspace{-2mm}
\end{figure}

Fig.~\ref{fig:heatmaps} shows the resulting signatures.
The onset boundary provides the clearest summary: it separates the box control from every structured support, and orders the structured supports among themselves, while a single threshold count at one $(m,\varepsilon)$ does not.
After whitening, the circle and ellipse have similar signatures, as expected.
This result illustrates the coordinate dependence discussed in \cref{sec:limitations}.

\section{Related work}\label{sec:related}

\paragraph{Empirical moment matrices and the Christoffel function.}
The matrix $\tfrac1n\Phi_m^\top\Phi_m$ with orthonormal polynomial features is the empirical moment matrix studied in the Christoffel--Darboux literature, where its spectrum is already used to read off algebraic properties of the support of a measure.
\citet{lasserre2019empirical} give a thresholding scheme for support approximation from finitely many moments with asymptotic guarantees, and establish consistency of the empirical Christoffel function at fixed degree.
\citet{pauwels2021data} use the eigendecomposition of the empirical moment matrix to identify algebraic properties of the support, including in machine-learning settings with data supported on singular sets.
\citet{lasserre2016sorting} use the inverse moment matrix for typicality and outlier scoring.
Perturbation of these objects has also been studied directly~\citep{beckermann2021perturbations}; see \citet{lasserre2022book} for a book-length treatment.
The moment matrix, its small-eigenvalue interpretation, and spectral thresholding are established components of this literature.

The closest prior work is \citet{rooshoefgeest2022cdtda}, who build a persistence module from the same empirical moment matrix and prove it stable in Wasserstein distance.
Unlike dAGP, they \emph{invert} the moment matrix to form the Christoffel polynomial $P_m(x)\!=\!\phi_m(x)^\top G_m^{-1}\phi_m(x)$.
They then compute persistent homology from its sublevel sets.
Their stability result controls bottleneck distance between persistence diagrams rather than the moment operator.
Both constructions use small-eigenvalue information: inversion weights polynomial directions by $1/\lambda_k(G_m)$ in the spatial function $P_m$, while dAGP counts eigenvalues at most $\varepsilon^2$ across degree and tolerance.
The distinction is between sublevel-set homology and spectral counts; dAGP requires no inversion and remains defined for singular $G_m$, the exact-relation case that their invertibility assumption excludes.
Their stability constant also depends on the data through $\|\log P_m\|_\infty$, which they interpret as a measure of algebraic degeneracy and describe as an undesirable dependence; ours depends only on the fixed feature map and on a convex set containing the supports.
dAGP differs in three specific ways: 
First, degree and tolerance remain \emph{explicit coordinates} of the representation.
Second, dAGP uses the monotone count $r^{\dAGP}_{m,\varepsilon}$ across this grid rather than a Christoffel function or inverse moment matrix at one degree.
Third, our stability result directly controls the moment operator and its ordered eigenvalues and yields threshold-count invariance under a spectral-margin condition.

\paragraph{Approximate vanishing ideals.}
Numerically stable border-basis and approximate-vanishing-ideal constructions, Vanishing Component Analysis, and later data-knotting or gradient-based refinements all recover polynomials that nearly vanish on sampled data~\citep{abbott2008stable,heldt2009approximate,livni2013vanishing,kera2018approximate,kera2020gradient}.
Generalised PCA models unions of subspaces with homogeneous equations~\citep{vidal2005gpca}, variety-learning work estimates defining equations and dimensions from finite samples~\citep{breiding2018learning}, and Hilbert functions have served as algebraic descriptors of subspace arrangements~\citep{yang2005hilbert}.
At a fixed degree, empirical AGP uses the same near-nullspace machinery as approximate-vanishing-ideal methods.
dAGP adds a population-level formulation without enforcing exact ideal generators or multiplicative closure.

\paragraph{Kernel and topological viewpoints.}
Kernel PCA eigendecomposes a nonlinear feature space and is normally read through dominant-variance directions~\citep{scholkopf1998kernelpca}; dAGP uses the lower end of the spectrum, where \eqref{eq:moment} gives an algebraic meaning.
Kernel mean embeddings represent distributions by feature expectations and may be characteristic for suitable kernels~\citep{smola2007hilbert,muandet2017kernel}.
Appendix~\ref{app:metrics} explains why the fixed-degree polynomial kernel is not characteristic.
Persistent homology organises homological features across a scale parameter and admits stable vector representations~\citep{cohensteiner2007stability,bubenik2015landscapes,adams2017images,carriere2020perslay,hensel2021survey}.
AGP borrows the persistence-style principle of tracking structure over parameters while targeting a different signal.
The two approaches may be complementary.
Evaluating this complementarity is future work.

\section{Limitations and future work}\label{sec:limitations}
The empirical evidence is limited to small synthetic datasets.
The experiments show that the signature is sensitive to controlled algebraic structure.
They also show that near-relation counts expose degree-organised onset structure more explicitly than $\lambda_{\min}$ alone.
However, these results do not establish predictive value on real data, and we have not compared the signature with AVI/VCA, kernel, or topological baselines.
Two further limitations are structural.
First, the feature dimension $M_m\!=\!\binom{d+m}{m}$ grows combinatorially, which restricts the current experiments to $d\!=\!2$ and small $m$.
Sketching or restricted feature sets may allow larger degrees or ambient dimensions.
Second, fixed-degree polynomial moments are not injective on $\cP(\Omega)$ (Appendix~\ref{app:metrics}), so distinct measures can share a signature.
Other limitations concern preprocessing, stability guarantees, coordinate dependence, and tolerance calibration.
Theorem~\ref{thm:w1} assumes a fixed degree, domain, and polynomial feature map.
Data-dependent centering or whitening therefore requires a separate stability analysis.
The certified radii for threshold stability (\cref{sec:stability}) are much smaller than the $W_1$ distances measured in our perturbation experiments.
The representation is coordinate-dependent, as illustrated in Fig.~\ref{fig:heatmaps}.
Tolerance calibration is empirical (\cref{sec:calibration}) but remains tied to a known noise model.

\paragraph{Future work.} A key next step is a matched-difficulty evaluation of the count grid, a fixed-length integer vector, against persistence images, kernel embeddings, and AVI/VCA (\cref{sec:samplesize}).
Two design choices require evaluation: whether to set the tolerance grid relative to a noise estimate (\cref{sec:calibration}), and whether to use the onset boundary instead of the full monotone count grid.
Once moments through degree $2m$ are available, the cost of assembling $G_m(\mu)$ is independent of $n$.
However, the $M_m \!\times\! M_m$ eigenproblem remains.
Sketching or restricted feature sets could reduce its cost.
Two theoretical questions remain open: whether a relative rather than an absolute perturbation bound can be obtained, and whether Theorem~\ref{thm:w1} extends through the whitening map.

{\small
\bibliographystyle{plainnat}
\bibliography{references}
}

\appendix

\section{Terminology and notation}\label{app:notation}
We provide a brief reference to the terminology and notation used, with an emphasis on algebraic notions that may be less familiar to some readers.
See \citet{cox2015ideals} for more on the affine and computational algebraic notions; see \citet{bochnak1998real} for more real algebraic geometry.

\subsection{Terminology}
We give terminology with context that aids the understanding of the paper.

\begin{description}[leftmargin=10pt, labelsep=0.5em, itemsep=0pt]

\item[Polynomial relation and defining equation.]
For a set $A\!\subseteq\!\R^d$, a polynomial $q\!\in\!\R[x_1,\dots,x_d]$ is a \emph{polynomial relation}\footnote{With noise, a polynomial will generally not vanish exactly, motivating the approximate relations used in this paper.} on $A$ if $q(x)\!=\!0$ for every $x\!\in\!A$.
A defining equation is a relation whose zero set, alone or together with other equations, determines the algebraic set of interest.

\item[Real algebraic set and algebraic curve.]
A real algebraic set\footnote{Some conventions use the word \emph{variety} for an irreducible algebraic set.
This paper needs no irreducibility assumption.} is the common zero set in $\R^d$ of a collection of real polynomials.
An algebraic curve is a 1-dimensional algebraic set.

\item[Vanishing ideal.]
The vanishing ideal of $A\!\subseteq\!\R^d$ is
$I(A)\coloneqq\{q\in\R[x_1,\dots,x_d]:q(x)\!=\!0 ~\forall x\!\in\!A\}$.
It is an ideal because $q_1\!+\!q_2\!\in\!I(A)$ whenever $q_1,q_2\!\in\! I(A)$, and $hq\!\in\! I(A)$ whenever $q\!\in\! I(A)$ and $h$ is any polynomial.
This multiplication property explains, for example, why a defining relation of degree $s$ produces additional relations at every degree above $s$.

\item[Zariski closure.]
The Zariski closure $\overline{A}^{\,Z}$ is the smallest real algebraic set containing $A$.
Equivalently, it is the common zero set of all polynomials that vanish on $A$.
Thus, $V\!=\!\overline{\operatorname{supp}\mu}^{\,Z}$ retains precisely the exact polynomial relations satisfied by the support.
The Zariski closure is distinct from the Euclidean closure.
If $\operatorname{supp}\mu$ contains a nonempty open subset of $\R^d$, then $V\!=\!\R^d$, which is one reason positive tolerances are essential here.

\item[Affine Hilbert function.]
For an algebraic set $V$, the affine Hilbert function used here is
\[
\operatorname{HF}_V(m)
~\coloneqq~
\dim\left(
S_{\leq m}/\bigl(I(V)\cap S_{\leq m}\bigr)
\right).
\]
It counts the number of linearly independent polynomial functions of degree at most $m$ remaining after exact relations on $V$ have been identified.
Here, ``dimension'' is algebraic, not homological.

\item[Moment matrix and near-null direction.]
The moment matrix $G_m(\mu)$ is the Gram matrix of the polynomial features through degree $m$ in $L^2(\mu)$.
The identity $c^\top G_m(\mu)c\!=\!\norm{f_c}_{L^2(\mu)}^2$ means that an eigenvector associated with a small eigenvalue represents a polynomial that is small on the distribution.
We call such a coefficient vector a \emph{near-null direction} or an \emph{approximate relation}.

\item[Approximate-vanishing cone.]
For $\varepsilon\!>\!0$, the set $\cC^{\dAGP}_{m,\varepsilon}(\mu)$ contains coefficient vectors whose polynomials are small in $L^2(\mu)$ relative to their coefficient norm.
The set is a cone because it is closed under scalar multiplication.
In general, it is neither a linear subspace nor an ideal because sums and polynomial multiples can exceed the fixed tolerance.
The degree-$4$ circle example in \cref{sec:signature} illustrates this fixed-tolerance failure of multiplicative closure: at $\varepsilon^2\!=\!5\!\times\!10^{-4}$, $18$ of $20$ draws have $r^{\dAGP}_{4,\varepsilon}\!=\!5$, rather than the noiseless prediction of six.
Lemma~\ref{lem:inertia} identifies the dimension of the cone's largest linear subspace.

\item[Relation onset.]
At a fixed tolerance, the relation-onset degree is the smallest $m$ for which $r^{\dAGP}_{m,\varepsilon}\!>\!0$.
Under noise it is a tolerance-dependent measure of low-degree algebraic approximability, not necessarily the minimal degree of an exact defining equation.

\item[Persistence-style.]
We use \emph{persistence-style} to mean tracking a nested quantity over polynomial degree and tolerance.
The resulting count is monotone in both parameters.
The term \emph{persistence-style} does not imply a persistence module, barcode, or persistence diagram.

\item[Christoffel--Darboux kernel.]
When $G_m(\mu)$ is invertible, the associated degree-$m$ kernel is
$K_m^\mu(x,y)\!=\!\phi_m(x)^\top G_m(\mu)^{-1}\phi_m(y)$.
Inversion weights spectral directions by reciprocal eigenvalues.
dAGP instead summarises the spectrum through near-relation counts across degree and tolerance, without requiring $G_m(\mu)$ to be invertible.
\end{description}

\subsection{Notation}
Table~\ref{tab:notation} collects the recurring symbols; local notation is defined where it appears.

{\small
\renewcommand{\arraystretch}{1.1}

\begin{longtable}{@{}p{0.331\linewidth}@{}p{0.669\linewidth}@{}}
\caption{Recurring notation used throughout the paper.}
\label{tab:notation}\\
Notation & Meaning\\
\hline\hline\noalign{\vskip 2pt}
\endfirsthead

\multicolumn{2}{l}{\small\textit{Table~\ref{tab:notation} continued}}\\
Notation & Meaning\\
\hline\hline\noalign{\vskip 2pt}
\endhead

\midrule
\multicolumn{2}{r}{\small\textit{Continued on the next page}}\\
\endfoot

\bottomrule
\endlastfoot

\multicolumn{2}{@{}l}{\textbf{Probability measures and data}}\\


$\Omega\!=\![-1,1]^d$ &
Normalised ambient domain.\\

$\cP(\Omega)$ &
Set of Borel probability measures supported on $\Omega$.\\



$K\subseteq\Omega$ &
Convex localisation set containing the supports of the measures being compared.\\



$Z=\{z_i\}_{i=1}^n$ &
Point cloud containing $n$ observations.\\

$\widehat\mu_Z
 =n^{-1}\sum_{i=1}^n\delta_{z_i}$ &
Equal-weight empirical measure associated with $Z$.\\

$\widehat\mu_n$ &
Empirical measure of an independent sample of size $n$ from $\mu$.\\

\multicolumn{2}{@{}l}{\textbf{Polynomial features}}\\


$\alpha=(\alpha_1,\dots,\alpha_d)$, $\lvert\alpha\rvert=\sum_j\alpha_j$ &
Multi-index and its total degree.\\

$S_{\leq m}=\R[x_1,\dots,x_d]_{\leq m}$ &
Vector space of real polynomials of total degree at most $m$.\\

$M_m=\binom{d+m}{m}$ &
Dimension of $S_{\leq m}$ and length of the degree-$m$ feature vector.\\

$P_k$, $\ell_k$ &
The degree-$k$ Legendre polynomial and its $L^2([-1,1])$-normalised version.\\

$p_\alpha$ &
Orthonormal multivariate polynomial basis element indexed by $\alpha$.\\

$\phi_m(x)=(p_\alpha(x))_{\lvert\alpha\rvert\leq m}$ &
Degree-$m$ polynomial feature map from $\Omega$ to $\R^{M_m}$.\\


$f_c=\sum_\alpha c_\alpha p_\alpha$ &
Polynomial represented by $c$, satisfying
$f_c(x)=\phi_m(x)^\top c$.\\

$D\phi_m(x)$ &
Jacobian matrix of the feature map.\\

\multicolumn{2}{@{}l}{\textbf{Moment matrices and spectral quantities}}\\

$G_m(\mu)$ &
Degree-$m$ moment matrix, equivalently the Gram matrix of the chosen polynomial features in $L^2(\mu)$, $\int\phi_m(x)\phi_m(x)^\top d\mu(x)$.\\

$\Phi_m(Z)$ &
Evaluation matrix whose $i$th row is $\phi_m(z_i)^\top$.\\

$\lambda_k(A)$, $\lambda_{\min}(A)$ &
Eigenvalues of a symmetric matrix $A$, ordered nondecreasingly, and its smallest eigenvalue.\\

$\sigma_k(A)$ &
Singular values of a matrix $A$.\\

$\varepsilon$ &
Population $L^2$ tolerance; the corresponding eigenvalue threshold is
$\varepsilon^2$.\\

$\tau$ &
Point-cloud singular-value tolerance.
Under Proposition~\ref{prop:empirical},
$\tau=\sqrt n\,\varepsilon$.\\

$\cC^{\dAGP}_{m,\varepsilon}(\mu)$ &
Cone of degree-$m$ coefficient vectors satisfying
$\norm{f_c}_{L^2(\mu)}\leq\varepsilon\norm{c}_2$.\\

$r^{\dAGP}_{m,\varepsilon}(\mu)$ &
Number of eigenvalues of $G_m(\mu)$ not exceeding $\varepsilon^2$; equivalently, the maximal dimension of a linear subspace contained in the approximate-vanishing cone.\\

$H^{\dAGP}_{m,\varepsilon}(\mu)\!=\!M_m-r^{\dAGP}_{m,\varepsilon}(\mu)$ &
Hilbert-style effective algebraic dimension, complementary to the near-relation count.\\

$\cC^{\AGP}_{m,\tau}(Z)$,
$r^{\AGP}_{m,\tau}(Z)$ &
Point-cloud approximate-vanishing cone and its spectral count.\\

\multicolumn{2}{@{}l}{\textbf{Algebraic quantities}}\\

$V=\overline{\operatorname{supp}\mu}^{\,Z}$ &
Zariski closure of the support of $\mu$.\\

$I(A)$, $I(V)$ &
Vanishing ideal of a set $A$, and in particular of the algebraic set $V$.\\

$\operatorname{HF}_V(m)$ &
Affine Hilbert function of $V$ through degree $m$.\\

\multicolumn{2}{@{}l}{\textbf{Metrics, norms, and stability constants}}\\



$\norm{f}_{L^2(\mu)}$ &
$L^2(\mu)$ norm
$\bigl(\int |f(x)|^2d\mu(x)\bigr)^{1/2}$.\\

$W_1(\mu,\nu)$ &
$1$-Wasserstein distance, using Euclidean transport cost on $\Omega$.\\


$B_{m,d}(K)$ &
Feature-size bound $\sup_{x\in K}\|\phi_m(x)\|_2$; $B_{m,d}=B_{m,d}(\Omega)$.\\

$L_{m,d}(K)$ &
Feature-Jacobian bound $\sup_{x\in K}\|D\phi_m(x)\|_2$; $L_{m,d}=L_{m,d}(\Omega)$.\\

$\Delta\!=\!2B_{m,d}(K)L_{m,d}(K)W_1(\mu,\nu)$ &
Upper bound on the moment-operator and ordered-eigenvalue displacement
in Corollary~\ref{cor:threshold}.\\


$\widetilde{k}_m(x,y) = \langle\phi_m(x),\phi_m(y)\rangle^2$ &
Lifted polynomial kernel whose kernel mean embedding is $G_m(\mu)$.\\

$\operatorname{MMD}_{\widetilde{k}_m}$ &
Maximum mean discrepancy induced by $\widetilde{k}_m$.\\







\multicolumn{2}{@{}l}{\textbf{Experiment}}\\

$r$ &
Radius used for the circle and quartic examples; not the spectral count $r^{\dAGP}_{m,\varepsilon}$.\\

$\sigma$ &
Standard deviation of the isotropic Gaussian perturbation; not the singular values $\sigma_k(A)$.\\


$r_0,a,k$ &
Base radius, radial-modulation amplitude, and angular frequency in $r(\theta)\!=\!r_0e^{a\cos(k\theta)}$; $k$ is distinct from the eigenvalue and singular-value indices in $\lambda_k(A)$ and $\sigma_k(A)$, and from the polynomial degree in $P_k$ and $\ell_k$.\\

$\lambda$ &
Ridge-whitening parameter, fixed at $\lambda\!=\!10^{-3}$ in \cref{sec:multishape}; 
not the eigenvalues $\lambda_k(A)$.
\end{longtable}
}

\section{Proofs}\label{app:proofs}

\subsection{Spectral index of the approximate-vanishing cone}\label{app:agp}

\begin{lemma}[Nonpositive-subspace dimension]\label{lem:inertia}
Let $A\!\in\!\R^{q\times q}$ be symmetric and let $\mathcal Q_A\!=\!\{c:c^\top A c\!\le\!0\}$.
If $n_-(A)$ and $n_0(A)$ denote the numbers of negative and zero eigenvalues of $A$, counted with multiplicity, then $\max\{\dim L:L\text{ a linear subspace},\,L\subseteq\mathcal Q_A\}=n_-(A)+n_0(A)$.
\end{lemma}
\begin{proof}
The direct sum of the negative and null eigenspaces lies in $\mathcal Q_A$, giving the lower bound.
Let $E_+$ be the positive eigenspace, of dimension $n_+(A)$.
If a subspace $L\subseteq\mathcal Q_A$ had dimension greater than $q-n_+(A)$, then $\dim L+\dim E_+>q$, so $L\cap E_+$ would contain a nonzero vector, which has strictly positive quadratic form, contradicting $L\subseteq\mathcal Q_A$.
\end{proof}

Setting $A\!=\!G_m(\mu)\!-\!\varepsilon^2I$ gives the identification of $r^{\dAGP}_{m,\varepsilon}$ with the maximal dimension of a linear subspace contained in the cone $\cC^{\dAGP}_{m,\varepsilon}$; setting $A\!=\!\Phi_m(Z)^\top\Phi_m(Z)\!-\!\tau^2I$, whose nonpositive eigenvalues are exactly those associated with $\sigma_k(\Phi_m(Z))\!\le\!\tau$, gives the point-cloud statement.
When $n\!<\!M_m$ the matrix $\Phi_m(Z)^\top\Phi_m(Z)$ has at least $M_m\!-\!n$ zero eigenvalues, and this is why the singular-value list must be zero-padded to length $M_m$.

\subsection{Nesting and monotonicity}\label{app:nesting}
Order the basis by total degree.
For $m\!\le\!m'$ and $c\!\in\!\R^{M_m}$ let $\iota_{m,m'}c$ be the zero-padding into $\R^{M_{m'}}$.
Zero-padding does not change the represented polynomial or the coefficient norm: $\Phi_{m'}(Z)\iota_{m,m'}c\!=\!\Phi_m(Z)c$ and $\norm{\iota_{m,m'}c}_2\!=\!\norm{c}_2$.
Thus $\iota_{m,m'}\cC^{\AGP}_{m,\tau}(Z)\!\subseteq\!\cC^{\AGP}_{m',\tau}(Z)$.
Nesting in $\tau$ follows directly from the definition, and the same argument with $\norm{f_c}_{L^2(\mu)}$ gives the dAGP statement.

Moreover, $G_m(\mu)$ is a principal submatrix of $G_{m+1}(\mu)$, so by Cauchy interlacing the number of eigenvalues strictly larger than any threshold $t$ cannot decrease from $m$ to $m+1$; hence $H^{\dAGP}_{m,\varepsilon}\!\le\! H^{\dAGP}_{m+1,\varepsilon}$ and $r^{\dAGP}_{m,\varepsilon}\!\le\! r^{\dAGP}_{m+1,\varepsilon}$, and analogously for the empirical Gram matrices.

\subsection{Proof of Proposition~\ref{prop:empirical}}
\begin{proof}
$G_m(\widehat\mu_Z)\!=\!\tfrac1n\sum_i\phi_m(z_i)\phi_m(z_i)^\top\!=\!\tfrac1n\Phi_m(Z)^\top\Phi_m(Z)$, and $\norm{f_c}^2_{L^2(\widehat\mu_Z)}\!=\!\tfrac1n\sum_if_c(z_i)^2\!=\!\tfrac1n\norm{\Phi_m(Z)c}_2^2$.
Thus $\norm{f_c}_{L^2(\widehat\mu_Z)}\!\le\!\varepsilon\norm{c}_2$ if and only if $\norm{\Phi_m(Z)c}_2\le\sqrt n\,\varepsilon\norm{c}_2$.
\end{proof}

\subsection{Proof of Theorem~\ref{thm:w1} and Corollary~\ref{cor:threshold}}
\begin{proof}
Let $A_m(x)\!=\!\phi_m(x)\phi_m(x)^\top$ and abbreviate $B=B_{m,d}(K)$, $L\!=\!L_{m,d}(K)$.
For $x,y\!\in\! K$, convexity of $K$ gives $\norm{\phi_m(x)-\phi_m(y)}_2\le L\norm{x-y}_2$ by the mean value inequality along the segment $[x,y]\subseteq K$, and
\begin{align*}
\norm{A_m(x)-A_m(y)}_2
&=\norm{\phi_m(x)(\phi_m(x)-\phi_m(y))^\top+(\phi_m(x)-\phi_m(y))\phi_m(y)^\top}_2
\\
&\le\bigl(\norm{\phi_m(x)}_2+\norm{\phi_m(y)}_2\bigr)\norm{\phi_m(x)-\phi_m(y)}_2
\le2BL\norm{x-y}_2.
\end{align*}
Any coupling $\pi\in\Pi(\mu,\nu)$ is supported in $K\times K$, so
\begin{align*}
\norm{G_m(\mu)-G_m(\nu)}_2
=\bigg\|\int\bigl(A_m(x)-A_m(y)\bigr)d\pi\bigg\|_2
&\le\int\norm{A_m(x)-A_m(y)}_2 d\pi
\\
&\le2BL\int\norm{x-y}_2 d\pi,
\end{align*}
and taking the infimum over couplings gives \eqref{eq:w1bound}.
The matrices are symmetric, so Weyl's perturbation inequality~\citep{bhatia1997matrix} gives the eigenvalue bound.
For the corollary, every ordered eigenvalue moves by at most $\Delta$, so under the stated margin condition none can cross $\varepsilon^2$; the number below the threshold, and hence its complement, are unchanged.
\end{proof}

Taking $K=\Omega$ recovers the unlocalised statement.
The constants are finite for each fixed $m$ because $\phi_m$ is a polynomial map and $K$ is compact.

\section{Metric and convergence details}\label{app:metrics}

\paragraph{Empirical convergence.}
Theorem~\ref{thm:w1} holds for any pair of measures, so taking $\nu\!=\!\widehat\mu_n$ and combining it with rates for the empirical Wasserstein distance~\citep{fournier2015rate} gives, for $\mu$ supported in $K$ and $d\!=\!2$,
\[
\E\|G_m(\widehat\mu_n)-G_m(\mu)\|_2
~\lesssim~
2B_{m,d}(K)L_{m,d}(K)
\dfrac{\log(1+n)}{\sqrt{n}}.
\]
Moreover $G_m(\widehat\mu_n)\!\to\!G_m(\mu)$ almost surely, since $W_1(\widehat\mu_n,\mu)\!\to\!0$ almost surely on a compact domain.
Thus the population object is the limit of the empirical one.
This is not a practical sample-complexity guarantee here.
The constant dominates the rate, and the resulting sample size required to certify threshold counts at the tolerances of \cref{sec:onset} is far larger than the sample sizes used here.

\paragraph{Why $W_1$, and what it does not give.}
$G_m(\mu)\!-\!G_m(\nu)$ depends only on the moments of $\mu\!-\!\nu$ up to order $2m$, so it is continuous under any metric that metrises weak convergence, and \eqref{eq:w1bound} also holds with $W_p$ for $p\!\ge\!1$ since $W_1\!\le\! W_p$.
Consider the lifted polynomial kernel
\[
\widetilde{k}_m(x,y)
~\coloneqq~
\langle\phi_m(x),\phi_m(y)\rangle^2.
\]
Its feature map may be taken as $x\mapsto\phi_m(x)\phi_m(x)^\top$ in the space of matrices equipped with the Frobenius inner product, so its kernel mean embedding is $G_m(\mu)$.
Consequently,
\[
\norm{G_m(\mu)-G_m(\nu)}_2
~\leq~
\norm{G_m(\mu)-G_m(\nu)}_{\mathrm F}
~=~
\operatorname{MMD}_{\widetilde{k}_m}(\mu,\nu).
\]
Thus $W_1$ provides a convenient and interpretable metric for the stability analysis.
The bound is one-directional: fixed-degree polynomial moments are not injective on $\cP(\Omega)$.
So $\mu\mapsto G_m(\mu)$ is Lipschitz but not bi-Lipschitz, and dAGP is structure-specific rather than a universal distribution representation.
Nor is dAGP itself a transport algorithm; $W_1$ supplies the geometry on $\cP(\Omega)$ under which the signature is stable.

\section{Experimental details}\label{app:experiments}
A single script regenerates every figure and table in this paper in under $15$ seconds on a standard laptop CPU; no GPU is used.
Exploratory experiments that did not reach the paper ran on the same hardware at comparable per-run cost.
All experiments use fixed NumPy seeds for reproducibility.
\cref{sec:perturbation}, \cref{sec:onset}, and~\cref{sec:samplesize} give each perturbation or point cloud its own child of a \texttt{SeedSequence}; each panel of Fig.~\ref{fig:calibration} averages over the same ten seeds at every plotted point.
The implementation uses NumPy, SciPy (BSD 3-Clause), and Matplotlib (PSF-based licence).
All point clouds are generated by the accompanying code.
No external dataset or pretrained model is used.

\paragraph{Constants.}
Table~\ref{tab:constants} reports numerical estimates obtained by evaluating $\norm{\phi_m(x)}_2$ and the largest singular value of the $M_m\!\times\!2$ Jacobian $D\phi_m(x)$ on a $401\!\times\!401$ grid over $[-1,1]^2$, restricted to $K$ in the localised column.
Replacing $2B_{m,2}L_{m,2}$ by a direct grid estimate of the Lipschitz constant of $A_m$ improves the estimated bound by under $4\%$ at $m=4$, so localisation accounts for most of the looseness in the bound, rather than the triangle inequality used in the proof.

\paragraph{Perturbation experiment (\cref{sec:perturbation}).}
$n\!=\!160$ latent points are drawn once on the circle of radius $0.72$ and held fixed.
At each $\sigma$, $12$ independent Gaussian perturbations are added and coordinates are clipped to $[-1,1]$; coordinatewise clipping is the Euclidean projection onto the cube and is $1$-Lipschitz, so it can only decrease $W_1$.
Exact $W_1$ between equal-weight empirical measures of equal size is the minimum-cost bipartite assignment under Euclidean cost.
Table~\ref{tab:noise} reports the summary.
Finite-sample Wasserstein convergence is governed by separate dimension- and moment-dependent theory~\citep{fournier2015rate}.

\begin{table}[h]
\centering
\caption{Mean (standard deviation) over 12 perturbations at each Gaussian noise level, $m=4$.}
\label{tab:noise}
\begin{tabular}{@{}rrrr@{}}
$\sigma$ & $W_1$ & $\|G_m(\widehat\mu_0)-G_m(\widehat\mu_\sigma)\|_2$ & $\max_k|\Delta\lambda_k|$\\
\hline\hline\noalign{\vskip 2pt}
0.005 & 0.0060 (0.0003) & 0.0028 (0.0004) & 0.0010 (0.0003)\\
0.010 & 0.0116 (0.0005) & 0.0058 (0.0007) & 0.0018 (0.0007)\\
0.020 & 0.0224 (0.0008) & 0.0130 (0.0028) & 0.0055 (0.0025)\\
0.040 & 0.0406 (0.0018) & 0.0266 (0.0048) & 0.0155 (0.0051)\\
0.070 & 0.0667 (0.0024) & 0.0573 (0.0063) & 0.0357 (0.0071)\\
0.100 & 0.0906 (0.0043) & 0.1003 (0.0170) & 0.0670 (0.0141)\\
0.140 & 0.1234 (0.0092) & 0.2019 (0.0471) & 0.1389 (0.0245)
\end{tabular}
\end{table}

\paragraph{Relation onset (\cref{sec:onset}--\cref{sec:calibration}).}
All four families use $n\!=\!300$ and 20 repetitions (10 per point in Fig.~\ref{fig:calibration}), without data-dependent whitening.
Noise $\sigma=0.01$ is added to the three curves but not to the box, which is already two-dimensional; adding the same noise to the box's uniform measure $\mu$ would raise $\lambda_{\min}(G_m(\mu))$ by at most $6.1\%$ for $m\le6$.
The quartic is parametrised by $x=r\,\mathrm{sgn}(\cos t)|\cos t|^{1/2}$, $y\!=\!r\,\mathrm{sgn}(\sin t)|\sin t|^{1/2}$, which satisfies $x^4+y^4\!=\!r^4$ exactly.
Unlike the circle parametrisation, the quartic parametrisation is not uniform in arclength.
Resampling the quartic uniformly by arclength changes $\lambda_{\min}$ by less than a factor of $1.3$ at every degree and leaves all onset degrees unchanged.
Thus the onset comparison is insensitive to this sampling-density difference.
The transcendental family is $r(\theta)\!=\!0.62\,e^{0.30\cos k\theta}$ with $k\!=\!5$ in Fig.~\ref{fig:onset} and $k\!=\!1,\dots,7$ in Fig.~\ref{fig:calibration}(b).
The accompanying script also records every family's counts at $\varepsilon^2\!\in\!\{5\!\times\!10^{-4},10^{-3},2\!\times\!10^{-3}\}$, including those quoted in \cref{sec:signature}, \cref{sec:onset}--\cref{sec:calibration}, and the box's smallest eigenvalues with and without noise.

\begin{figure}[h]
\centering
\includegraphics[width=0.6\linewidth]{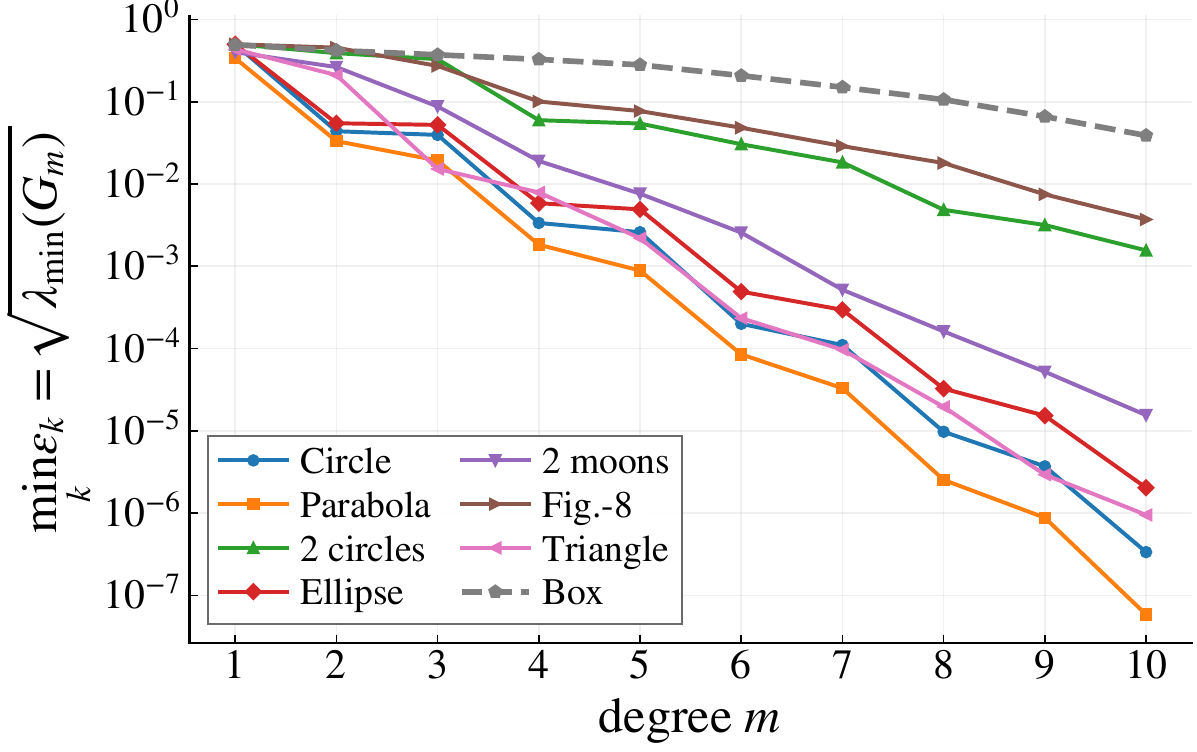}
\caption{Smallest population-scale singular value $\min_k\varepsilon_k\!=\!\sqrt{\lambda_{\min}(G_m)}$ across the eight synthetic families.
From $m\!=\!3$ onwards, the uniform-box control stays farthest from singular.
Several algebraic supports decay much earlier. 
Their ordering does not follow defining degree, which motivates the count-based analysis (rather than the extreme value) in~\cref{sec:onset}.
}
\label{fig:minsv}
\end{figure}

\paragraph{Multi-shape diagnostic (\cref{sec:multishape}).}
We use eight families with $n\!=\!250$ and a fixed seed: a unit circle, $y\!=\!x^2$, two separated circles, an ellipse, two moons, a figure-eight $(\sin t,0.55\sin2t)$, a triangle boundary, and a uniform sample from $[-1,1]^2$.
We add small isotropic Gaussian perturbations to the structured supports.
Preprocessing is centering, ridge whitening with $\lambda\!=\!10^{-3}$, and coordinatewise rescaling into $[-1,1]^2$, followed by the tensor-product Legendre basis specified in \cref{sec:signature}, for $m\!=\!1,\dots,10$.
Fig.~\ref{fig:minsv} reports the smallest population-scale singular value across the same families.

\paragraph{Sample-size sensitivity (\cref{sec:samplesize}).}
For each $n\!\in\!\{50,100,200,300,500\}$ and each of the four relation-onset families, we generate $100$ independent samples using the same family parameters and noise settings as in the relation-onset experiment.
Reference onsets are computed from independent samples of size $20{,}000$ at $\varepsilon^2\!=\!5\!\times\!10^{-4}$.
The dAGP vector contains counts for $m\!=\!1,\ldots,6$ at ten squared tolerances logarithmically spaced from $10^{-4}$ to $10^{-2.3}$; the baselines are $(\log_{10}\lambda_{\min}(G_m))_{m=1}^6$ and the complete ordered log-spectrum of $G_6$.
Each cross-validation fold holds out one repetition---one cloud per family---and fits feature standardisation and the four nearest centroids on the other $99$ repetitions.
The count vector thus has $60$ coordinates, the minimum-eigenvalue profile $6$, and the full spectrum $M_6\!=\!28$.

\end{document}